\documentclass[conference]{IEEEtran}
\IEEEoverridecommandlockouts
\usepackage{cite}
\usepackage{amsmath,amssymb,amsfonts}
\usepackage{algorithm}
\usepackage{graphicx}
\usepackage{textcomp}
\usepackage{xcolor}
\usepackage{algpseudocode}
\algnewcommand\algorithmicforeach{\textbf{for each:}}
\algnewcommand\ForEach{\item[ \algorithmicforeach]}

\usepackage[font=scriptsize]{subcaption}
\usepackage{multicol}
 \usepackage{relsize}

\def\BibTeX{{\rm B\kern-.05em{\sc i\kern-.025em b}\kern-.08em
    T\kern-.1667em\lower.7ex\hbox{E}\kern-.125emX}}
\begin{document}

\title{Network Elastic Net for Identifying Smoking \\ specific gene expression for lung cancer\\
}

\author{\IEEEauthorblockN{Avinash Barnwal}
\IEEEauthorblockA{
\textit{Stony Brook University}\\
Stony Brook, USA \\
barnwal.avinash@stonybrook.edu}
}

\maketitle

\begin{abstract}
Survival month for non-small lung cancer patients depend upon which stage of lung cancer is present. Our aim is to identify smoking specific gene expression biomarkers in prognosis of lung cancer patients. In this paper, we introduce the network elastic net, a generalization of network lasso that allows for simultaneous clustering and regression on graphs. In network elastic net, we consider similar patients based on smoking cigarettes per year to form the network. We then further find the suitable cluster among patients based on coefficients of genes having different survival month structures and showed the efficacy of the clusters using stage enrichment. This can be used to identify the stage of cancer using gene expression and smoking behavior of patients without doing any tests.
\end{abstract}


\section{Introduction}
One of the key challenge in molecular medicine is to personalize the feature selection for each data point on sample space. This can be treated as local feature selection and prediction problem. Recently, network lasso has been introduced where coefficient difference is penalized with l2 norm given graph structure.  It has already been observed that Elastic net works better than lasso with real world data for multivariate regression problems. We have introduced elastic net norm for network regression and tested the results for Survival months for non-small lung cancer patients for early stage cancer patients.

\section{Formulation and Methodology}

We are focusing on  optimization problems posed on graphs. Consider the following problem on a graph G = (V,E), where V is the vertex set and E is the set of edges:
\begin{equation}
minimize \sum_{i\epsilon v}f_i(x_i)+ \sum_{(j,k)\epsilon e}g_{jk}(x_j,x_k)
\end{equation}

The variables are ${x_1,...,x_n \epsilon R^p}$ , where ${n=|V|}$.(The total number of scalar variables is $np$.)  Here $x_i \epsilon R^p$ is the variable at node $i,f_i:R^p \rightarrow R \cup {\infty}$ is the cost function at node i, and $g_{jk}:R^p \times R^p \rightarrow R \cup {\infty}$ is the cost function associated with edge$(j,k)$. We use extended (infinite) values off i and $g_{jk}$ to describe constraints on the variables, or pairs of variables across an edge,respectively.Our focus will be on the special case in which the fiare convex, and ${g_{jk}(x_j,x_k) = \lambda_1w_{jk}||x_j-x_k||_2 + \lambda_2w_{jk}||x_j-x_k||}$, with ${\lambda \geq 0}$ and user-defined weights ${w_{jk} \geq 0}$:
\begin{equation}
\resizebox{0.5\textwidth}{!}{$ minimize \sum_{i\epsilon v}f_i(x_i)+\lambda_1\sum_{(j,k)\epsilon e}w_{jk}||x_j-x_k||_2 + \lambda_2\sum_{(j,k)\epsilon e}w_{jk}||x_j-x_k|| $}
\end{equation}

(2)The edge objectives penalize differences between the variables at adjacent nodes, where the edge between nodes i and j has combination of $l_2$ and $l_1$ norm having weights $\lambda_1w_{ij}$ and $\lambda_2w_{ij}$.   $w_{ij}$ can be considered as the graph property having similar nodes with more penalization, and  $\lambda_1$ and $\lambda_2$ as an overall parameter that scales the edge objectives relative to the node objectives.  We call problem (2) the network regression with elastic net problem, since the edge cost is a sum of $l1$ and $l2$ norms of differences of the adjacent edge variables. Hallac et. al has showed that network lasso is a convex optimization problem,and we have added a convex function part to it which maintains property of convex optimization problem. Here , problem requires scalable solution with large nodes and variables. Our proposed regularizer can be solved by a general alternating direction method of multipliers (ADMM) based solver.

ADMM consists the following steps (for each iteration k)

\begin{equation}
\begin{aligned}
x^{k+1} = \operatorname{arg\,min}_x L_p(x,z^k,u^k) \\
z^{k+1} = \operatorname{arg\,min}_z L_p(x^{k+1},z,u^k)\\
u^{k+1} = u^k + (x^{k+1} - z^{k+1})
\end{aligned}
\end{equation}

Each step can be described as x - update, z - update, and u - update.
As there will not be any changes in x-update and u-update compared to network lasso.But z-update is having extra $l1$-norm which is solved using soft-thresholding operator.

Following is the algorithm for ADMM Step

\begin{algorithm}[H] \caption{ADMM Steps}

\begin{algorithmic}[1]

\Repeat\\
$x^{k+1}_i= \underset{x_i}{\operatorname{argmin}}\big(f_i(x_i) + \mathlarger{\mathlarger{\sum}}_{j \in N(i)}(\rho/2)\parallel x_i - z^k_{ij}+u^k_{ij} \parallel^2_2\big)$ \\
$a=x_i^{k+1}+u_{ij}^k,b=x_j^{k+1}+u_{ji}^k$\\
$c_1=\lambda_1*(1-\alpha)*w_{ij},c_2=\lambda_1*(\alpha)*w_{ij}$\\
$\mu_1=||\rho*(a-b)-2*c_2||_2/\rho-2*c_1/\rho $ \\
$\mu_2=||\rho*(a-b)+2*c_2)/\rho-2*c_1/\rho$ \\
$\gamma_1=\mu_1*c_2/(2*c_1+\mu_1*\rho)$ \\
$\gamma_2=\mu_2*c_2/(2*c_1+\mu_2*\rho)$ \\
$\theta_1=0.5+\mu_1*\rho/(4*c_1+2*\mu_1*\rho),\theta_2=0.5+\mu_2*\rho/(4*c_1+2*\mu_2*\rho)$
%\EndFor
\Until {$||r^k||_2 \leq \epsilon^{pri}; ||s^k||_2 \leq \epsilon^{dual} $}

\end{algorithmic}
\end{algorithm}

Following is the algorithm for Regularization path

\begin{algorithm}[H] \caption{Regularization Steps}
\begin{algorithmic}[2]
\For {{\scriptsize Weight Type} $\in$ {\scriptsize \{Euclidean,Correlation,Diffusion Map\}}}
	\For {$\alpha$ $\in$ [0,1]}
		\State {\scriptsize \textbf{initialize} Solve for $x^*,u^*,z^*$ at $\lambda= 0.$}
        \Repeat
        \State set $\lambda:=\gamma\lambda;\gamma \ge 1$;
        \State \parbox[t]{\dimexpr\linewidth-\algorithmicindent}{Use Algorithm 
        1 to solve \\
        for $x^*(\lambda),u^*(\lambda),z^*(\lambda)$.\strut}
        \Until {$x^*(\lambda)$ = $x^*(\lambda_{previous})$}
        \State \textbf{return} $x^*(\lambda)$ for $\lambda$ from 0 to $\lambda_{critical}$
	\EndFor
\EndFor
\end{algorithmic}
\end{algorithm}

\section{Experiments}
In this section , we first illustrate our proposed method on synthetic data and then we perform the survival month prediction and clustering of survival months using coefficients of predictors and then for TCGA datasets, predicting survival months for LUAD lung cancers using gene expression variables.

\subsection{Synthetic experiments}

We have used high dimensional synthetic data having clustered and orthogonal coefficients .
First, we have generated the predictors such that $x_{ij} \sim Unif(-1,1) $ , j = 1,...,10 and i = 1,...,100 and $e_i \sim N(0,1) $.
and corresponding response variable have been generated using below model :-

\begin{equation}
y_i =  2*x_{i2} + 3*x_{i3} + 2*x_{i4}  + 0.1*e_i, i = 1,..,33
\end{equation}
\begin{equation}
y_i =  4*x_{i3} - 6*x_{i4} - 5*x_{i5}  + 0.1*e_i, i = 34,...,66
\end{equation}
\begin{equation}
y_i = -5*x_{i4} + 6*x_{i5} + 3*x_{i6}  + 0.1*e_i , i = 67,...,100
\end{equation}
        
Lets consider the first equation corresponds to first cluster , second equation corresponds to second cluster and third equation corresponds to third cluster. We have one more requirement here, link function , its very important to create the link function such that first cluster has high density among themselves but low density compared to other cluster , same holds for other cluster.

We have generated the link function using following steps:-\\
1. Create Sparse matrix R $ \in \{0,1\}^{100x100}$ with density 1\%. \\
2. Create Dense matrix  R $ \in \{0,1\}^{33x33}$ with density 95\%. \\
3. Create Dense matrix  R $ \in \{0,1\}^{34x34}$ with density 95\%. \\
4. Replace corresponding dense matrix in the sparse matrix such that first cluster has high density among themselves but low density compared to other clusters and similar properties for other two clusters. 

We experimentally set the regularization parameter for the proposed method to $\alpha$ $\in$ \{0,0.2,0.4,0.6,0.8,1\} and $\lambda$ = 1.12 . For the network Lasso, $\alpha$ = 0. Moreover, in we can also regularize the coefficients as well. First part of charts show estimated coefficients using our proposed method without coefficients regularization and second part of charts show estimated coefficients using our proposed method with coefficients regularization. We have used $\mu$ = 0.1 for coefficient regularization.

Clearly , we can see that estimated coefficients have been recovered for each $\alpha$ and for regularized coefficients $\alpha$ = 1 works best.

\begin{figure}[htbp]
\begin{multicols}{2}
\begin{subfigure}{0.22\textwidth}
\includegraphics[width=\linewidth]{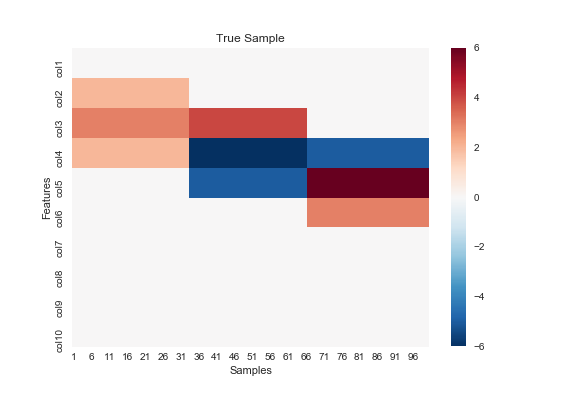}
\caption{True pattern.} \label{fig:1a}
\end{subfigure}
\begin{subfigure}{0.22\textwidth}
\includegraphics[width=\linewidth]{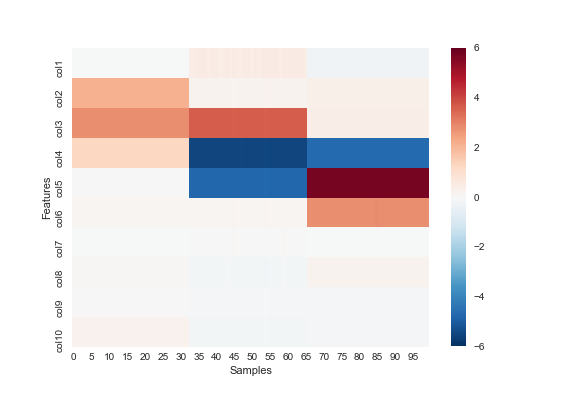}
\caption{$\lambda$ =1.12 , $\alpha$ = 0 and $\mu$ = 0.} \label{fig:1b}
\end{subfigure}
\begin{subfigure}{0.22\textwidth}
\includegraphics[width=\linewidth]{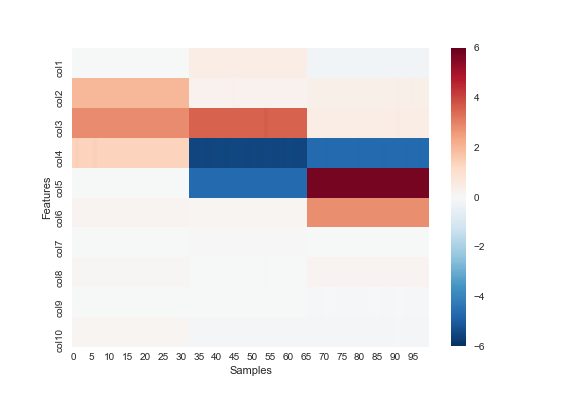}
\caption{$\lambda$ =1.12 , $\alpha$ = 0.2 and $\mu$ = 0.} \label{fig:1c}
\end{subfigure}
%\end{multicols}
\begin{subfigure}{0.22\textwidth}
\includegraphics[width=\linewidth]{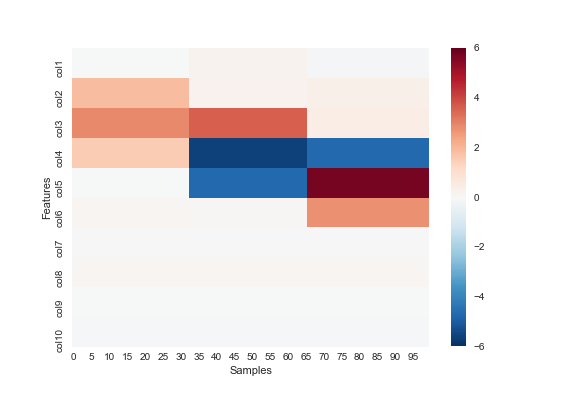}
\caption{$\lambda$ =1.12 , $\alpha$ = 0.4 and $\mu$ = 0.} \label{fig:1d}
\end{subfigure}
\begin{subfigure}{0.22\textwidth}
\includegraphics[width=\linewidth]{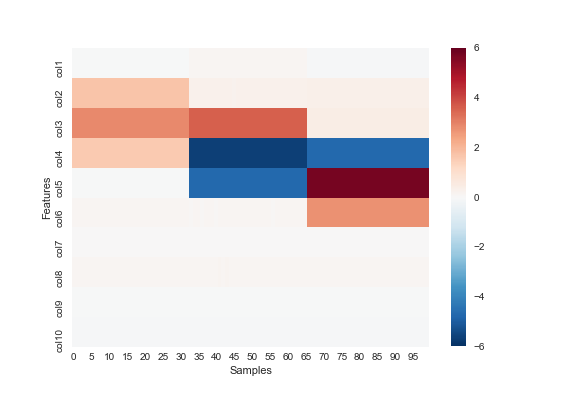}
\caption{$\lambda$ =1.12 , $\alpha$ = 0.6 and $\mu$ = 0.} \label{fig:1e}
\end{subfigure}
\begin{subfigure}{0.22\textwidth}
\includegraphics[width=\linewidth]{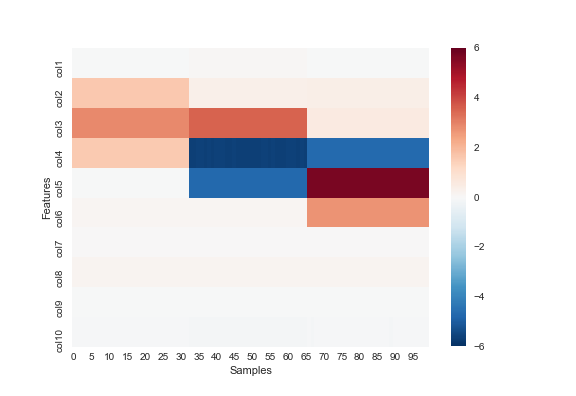}
\caption{$\lambda$ =1.12 , $\alpha$ = 0.8 and $\mu$ = 0.} \label{fig:1f}
\end{subfigure}
\begin{subfigure}{0.22\textwidth}
\includegraphics[width=\linewidth]{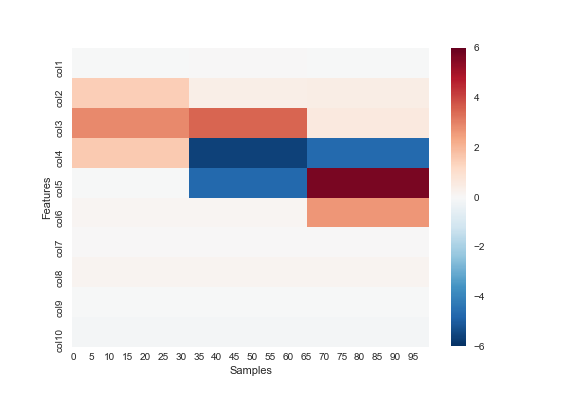}
\caption{$\lambda$ =1.12 , $\alpha$ = 1 and $\mu$ = 0.} \label{fig:1g}
\end{subfigure}
\end{multicols}
\caption{Simulation of Network Elastic net} 
\label{fig:1}
\end{figure}

%\vfill
%\clearpage

\subsection{Predictions of Survival Months for LUAD Cancer}

We have developed accelerated failure time-framework for modeling survival times. Accelerated failure-time is censored regression where inputs and responses are adjusted based on censored events.

\cite{aft} discusses Accelerated Failure Time being weighted least square regression, where, \\ 
$Y_{(1)} \leq ... \leq Y_{(n)}$ - order statistics of $Y_i's$. \\ $\delta_{(1)},...,\delta_{(n)}$ - associated censoring indicators. \\ 
$X_{(1)},...,X_{(n)}$ - associated covariates.
and it is further mean adjusted leading to below equations.

Denote 
\begin{equation}
X^{*}_{(i)} = {(nw_i)}^{\frac{1}{2}}(X_{(i)}-\overline{X}_w)
\end{equation}
\begin{equation}
Y^{*}_{(i)} = {(nw_i)}^{\frac{1}{2}}(Y_{(i)}-\overline{Y}_w)
\end{equation}
Therefore weighted loss function is
\begin{equation}
L = \sum\limits_{i=1}^n{(Y^{*}_{(i)}-X^{*}_{(i)}\beta)}^2
\end{equation}

For Network Elastic-Net 

\begin{align*}
L = \sum\limits_{i=1}^n{(Y^{*}_{(i)}-X^{*}_{(i)}\beta)}^2 +  \lambda(1-\alpha)\sum\limits_{(j,k)\epsilon E}w_{jk}{\| \beta_j-\beta_k \|}_2 + \\
 \lambda\alpha\sum\limits_{(j,k)\epsilon E}w_{jk}{\| \beta_j-\beta_k \|}_1
\end{align*}

We perform the survival month prediction using gene expression data. We have used TCGA dataset having lung cancer patients with clinical variables and corresponding gene expression data using accelerated failure time modeling. There are 325 patients in LUAD lung cancer and 423 patients in LUSC lung cancer.

We then divided the data into 80:20 to train the model. Weight measures based on number of smoking cigarettes per year.If $w_i$ is number of smoking cigarettes per year for $patient_i$ and similarly $w_j$ corresponding for $patient_j$.

For weight measure
\begin{equation}
    w_{ij}=
    \begin{cases}
      \frac{1}{w_i-w_j}, & \text{if}\ i \neq j \\
      0, & \text{otherwise}
    \end{cases}
  \end{equation}  
 
We have kept only 100 top correlated genes with survival months among 19196 as predictors for each cancer types and lets call it $gene_1 , \ gene_2 , ... , \ gene_{100}$. 

For each patient ,we solve for $x_i=[a_{i1} \ a_{i2} ... \ a_{i100}]^T$,which gives us the coefficients of the regressors. The survival months estimate is given by 

$y_i=a_{i1} \cdot gene_1 \ + \ a_{i2} \cdot gene_2 \ + \ .... \ + \ a_{i3} \cdot gene_100 + c_i $ 

,where the constant offset $c_i$ is the “baseline”. The objective function for each patient  then becomes $f_i=\parallel{Survival \ Month}_i \ - \ y_i \parallel_2^2  $ where , $Survival \ Month_i$ is the actual survival month for $patient_i$ .
$\\$

\noindent
Robust Cross-Validation is important step to find the hyper-parameters. We have split the data into 80:20 where Training data is 80\% and Testing data is 20\%. For performance measurement, \cite{aft} uses AIC.

\begin{equation}
    AIC-Score = nlog(CV Score) + 2K
\end{equation}

where CV Score is the loss for test data and K is corresponding non-zero coefficients. We calculated the best $\lambda$ based on that.

To predict the survival month on the test set, we connect each new patient to  the 5 nearest patient based on weight measures defined above.  We then infer the value of $x_j$ at for $patient_j$ by solving problem , and we use this value to estimate the new survival month for $patient_j$.

We  solve  for $x_j$ 
\begin{equation}
min\sum_{k\epsilon N(j)}w_{jk}\parallel x_j-x_k \parallel_2 
\end{equation}

The inference for new patient is being used by \cite{Network_Lasso} which is a weber problem.

Results for LUAD Cancer type:-

\begin{figure}[htbp]
\begin{multicols}{2}
\begin{subfigure}{0.22\textwidth}
\includegraphics[width=\linewidth]{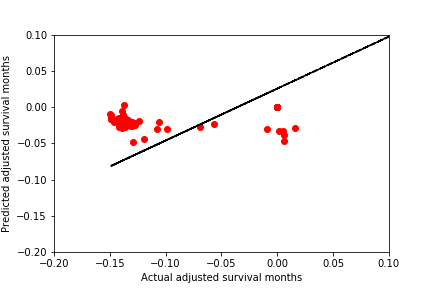}
\caption{Actual vs Predicted normalized survival months} \label{fig:1a}
\end{subfigure}
\begin{subfigure}{0.22\textwidth}
\includegraphics[width=\linewidth]{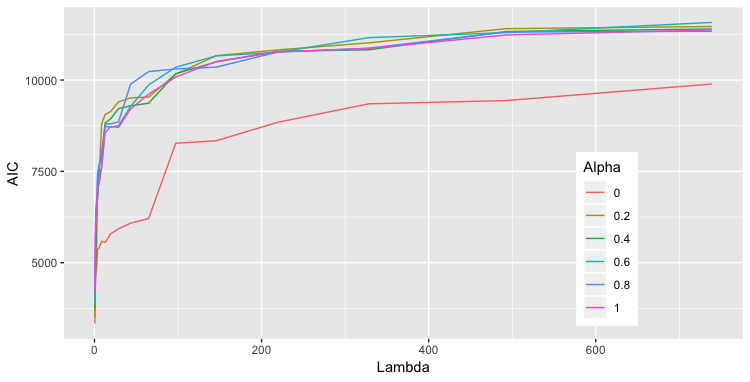}
\caption{Regularization Path} \label{fig:1c}
\end{subfigure}
\begin{subfigure}{0.3\textwidth}
\includegraphics[width=\linewidth]{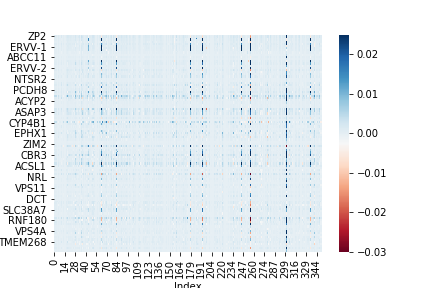}
\caption{Coefficients Heat Map} \label{fig:1d}
\end{subfigure}
\end{multicols}
\end{figure}

\subsection{Clustering of Patients - LUAD Cancer Patients}
Next, we have clustered the patients in training dataset based on gene expression coefficient calculated. Characteristics of clusters can further be validated based on external variables such as Cancer Stage, smoking stages and Km-plot. We have used smoking stage to identify the cluster characters. Cancer stage is ranged from 1 to 4. 	   

\begin{center}
\begin{tabular}{p{1.2cm} p{1.2cm} p{1.2cm} p{1.2cm} p{1.2cm}}
 \hline
 \multicolumn{5}{ c }{\textbf{Count of Stage of Cancer vs Cluster}} \\
 \hline
 Cluster & Stage 1 & Stage 2 & Stage 3 & Stage 4\\
 \hline
 1  &  53	& 26	& 12 & 4 \\
 2  &  3	& 2	    & 1	 & 0   \\
 3  &  36	& 13    & 13 & 4 \\
 4  &  45	& 20    & 17 & 6 \\
 5  &  13	& 6	    & 6	 & 1 \\
 \hline
\end{tabular}
\end{center}
$\\$
To identify which cluster corresponds to which stage , we have used similar to gene enrichment strategy where for each cluster and each stage we have calculated corresponding P-value  and if P-Value is smaller than <0.05 then cluster represents that stage.
\begin{center}
\begin{tabular}{p{1.2cm} p{1.2cm} p{1.2cm} p{1.2cm} p{1.2cm}}
 \hline
 \multicolumn{5}{ c }{\textbf{Significance of Stage of Cancer vs Cluster}} \\
 \hline
 Cluster & Stage 1 & Stage 2 & Stage 3 & Stage 4\\
 \hline
 1  &   0.00 &	 0.00 &	 0.86 &	 0.77 \\
 2  &   0.42 &	 0.32 &	 1.00 &	 1.00 \\
 3  &   0.00 &	 0.59 &	 0.11 &	 0.49 \\
 4  &   0.00 &	 0.11 &	 0.04 &	 0.13 \\
 5  &   0.14 &	 0.60 &	 0.24 &	 1.00 \\
 \hline
\end{tabular}
\end{center}

\section{Conclusion}
\begin{itemize}
\itemsep0em
\item $\lambda$ = 0.5 and $\alpha$ = 0 shows least \textbf{AIC} - 3331.
\item Correlation between Actual adjusted survival months and predicted adjusted survival months is 0.77. 
\item 31 Significant Genes having norms greater than 0.05.
\item Cluster 1 belongs to Stage 1 or Stage 2, Cluster 3 belongs to Stage 1 and Cluster 4 belongs to Stage 1 or Stage 3.
\end{itemize}


\begin{thebibliography}{00}
\bibitem{Network_Lasso} David Hallac, Jure Leskovec, Stephen Boyd {\em Network Lasso: Clustering and Optimization in LargeGraphs}. In KDD,2015.
\bibitem{aft} Jian  Huang,  Shuangge  Ma,  and  Huiliang  Xie.   Regularized  estimationin the accelerated failure time model with high-dimensional covariates.Biometrics, 62(3):813–820, 2006
\bibitem{Local_Lasso}Makoto Yamada, Koh Takeuchi, Tomoharu Iwata, John Shawe-Taylor, Samuel Kaski {\em Localized Lasso for High-Dimensional Regression}. arXiv:1603.06743, 2016
 
\bibitem{norman} Robert. Tibshirani{\em Regression shrinkage and selection via the Lasso}. Journal of the Royal Statistical Society,Series B, 58(1):267–288, 1996

\bibitem{Elasticnet} Hui Zou, Trevor Hastie {\em Regularization and variable selection via the elastic net}. Journal of the Royal Statistical Society,Series B, 67(2):301–320, 2005.

\bibitem{fo}Rie Kubota Ando and Tong Zhang{\em A frame work for learning predictive structures from multiple tasks and unlabeled data}. Journal of Machine Learning Research,6(Nov):1817–1853,2005

\bibitem{ADMM}S. Boyd, N. Parikh, E. Chu, B. Peleato, and J.Eckstein {\em Distributed optimization and statistical learning via the alternating direction method of multipliers}. Foundations and Trends in Machine Learning, 3:1–122, 2011
 
\bibitem{opti}S. Boyd and L. Vandenberghe {\em Convex Optimization} .Cambridge University Press, 2004

\bibitem{snap} J. Leskovec and R. Sosic.  {\em Snap.py: SNAP for Python} .http://snap.stanford.edu, 2014

\end{thebibliography}
\end{document}